\documentclass{article}
\pdfoutput = 1

\usepackage{ijcai24}

\usepackage{booktabs}
\usepackage{color}
\usepackage{times}
\usepackage{soul}
\usepackage{url}
\usepackage[hidelinks]{hyperref}
\usepackage[utf8]{inputenc}
\usepackage[small]{caption}
\usepackage{graphicx}
\usepackage{amsmath}
\usepackage{amsthm}
\usepackage{booktabs}
\usepackage{algorithm}
\usepackage{algorithmic}
\usepackage{comment}
\usepackage[switch]{lineno}
\usepackage{ mathrsfs }
\usepackage{enumerate}
\usepackage{subcaption}
\usepackage[inkscapeformat=png]{svg}

\title{Synthetic Data Aided Federated Learning Using Foundation Models}

\author{
Fatima Abacha$^1$ \and
 Sin G. Teo$^2$
 \and
 Lucas C. Cordeiro$^1$ and
Mustafa A. Mustafa$^{1,3}$
\affiliations
$^1$Department of Computer Science, The University of Manchester, UK\\
$^2$Institute for Infocomm Research, A*STAR, Singapore\\
$^3$COSIC, KU Leuven, Belgium\\
\emails
\{fatima.abacha,lucas.cordeiro,mustafa.mustafa\}@manchester.ac.uk;
teosg@i2r.a-star.edu.sg
}
\begin{document}

\maketitle

\begin{abstract}
%Federated Learning (FL) is a machine learning technique that enables multiple parties to %come  and 
%train together a model without sharing their data. %FL protects against data leakage as the data used in training the model is kept with the participating clients. 
% \textcolor{blue}{[MM: This is too much background info for the abstract. Remove it, or summarise it in one sentence.]}
In heterogeneous scenarios where the data distribution amongst the Federated Learning (FL) participants is Non-Independent and Identically distributed (Non-IID), FL suffers from the well-known problem of data heterogeneity. This leads the performance of FL to be significantly degraded, as the global model tends to struggle to converge. 
% \textcolor{blue}{[MM: Highlight even more the significance of the problem, by using words like `well-known limitation/issue/problem/etc.']}
To solve this problem, we propose Differentially Private Synthetic Data Aided Federated Learning Using Foundation Models (DPSDA-FL) -- a novel data augmentation strategy that aids in homogenizing the local data present on the clients’ side. DPSDA-FL improves the training of the local models by leveraging differentially private synthetic data generated from foundation models.
% \textcolor{blue}{[MM: This sentence is too long, split it into two. Also, highlight the originality of your approach; use words like `novel/innovative`]}
We demonstrate the effectiveness of our approach by evaluating it on the benchmark image dataset: CIFAR-10. Our experimental results have shown that DPSDA-FL can improve class recall and classification accuracy of the global model by up to 26\% and 9\%, respectively, in FL with Non-IID issues. 
% \textcolor{blue}{[MM: Add some concrete numbers here. Also mention the data sets used in the experiments. Here you need to say how rigorous your evaluation was.]}
\end{abstract}

\section{Introduction}

Federated Learning (FL) enables the training of a  machine learning  model by several parties without sharing their data with each other~\cite{mcmahan_communication-efcient_2017}. The training process is orchestrated by a third party, which is usually a central server. In FL, each client uses its private data to train its own model known as the local model, while the server uses an aggregation algorithm to construct a global model from the local models. The entire process runs for several iterations until a global model with the desired performance is achieved \cite{shahid_communication_2021}. This global model is then broadcast to all the clients to use it for inference on their test dataset. 

FL provides protection against data leakage as the private training data of each client is not disclosed to any other party. It can facilitate collaboration between institutions that deal with sensitive data, such as health and financial data \cite{aouedi_handling_2022}. Regulations such as the General Data Protection Regulation (GDPR) and Health Insurance Portability and Accountability Act (HIPAA) control how sensitive data are stored and shared within and between institutions in order to protect the privacy of the individuals whose data is captured \cite{zhou_privacy-preserving_2020}. FL can aid collaborators in adhering to these regulations, as no data is shared between the clients during the training or inference process. 

However, FL comes with its own challenges, as studies have shown that the global model struggles to converge when the data distribution amongst the clients is statistically heterogeneous \cite{zhao_federated_2018}, \cite{li_federated_2020}. This implies that the data distribution is Non-Independent and Identically distributed (Non-IID). A client may hold data from some classes, but not all classes present in the global dataset or clients could hold data for all classes but in different quantity. This statistical heterogeneity of local data could result in each local model being very different from other local models, leading to a global model that performs at a subpar level \cite{li_federated_2022}. Also, when clients train their local model on data that does not contain certain classes from the global set or only a few samples from specific classes, the models are likely to be biased towards those underrepresented groups \cite{hao_towards_2021}. This could lead to devastating consequences when these models are deployed in safety-critical situations such as healthcare and finance. 

The presence of biases could also disincentivize clients from participating in FL collaboration as they would lose trust in the system. For instance, imagine a collaboration between pharmaceutical companies training a model to determine the effectiveness of several drugs on an ailment and having the drug from one company consistently being predicted as the most effective because they provide more data as a result of conducting more experiments than the others \cite{rance_attacks_2023}. Data heterogeneity, as such, is a challenge that needs to be addressed to obtain trustworthy FL models.

 Some existing work~\cite{zhao_federated_2018} have proposed a global data sharing strategy to tackle the challenge of FL with Non-IID data. The server is posited to have a uniformly distributed dataset in its possession. This global data is then shared amongst the clients to harmonize their data distribution to alleviate the impact of data heterogeneity. Other approaches such as FedProx \cite{li_federated_2020} introduce a regularization term to the local model loss function on the client side, this mitigates the effect of data heterogeneity and enhances the convergence of the global model. The work of \cite{li_federated_2022} employs Generative Adversarial Networks (GANs) to produce synthetic data to solve the problem of Non-IID data in FL. The synthetic data is then augmented with the local data of the clients to improve the stability of the FL training process. While these proposed methods have improved the performance of FL with data heterogeneity, they are constrained by certain limitations. The assumption that the server possesses a uniformly distributed global dataset in \cite{zhao_federated_2018} is impractical in real-world FL scenarios. In contrast, the regularization technique employed by FedProx \cite{li_federated_2020} is not effective in extreme cases of data heterogeneity. On the other hand, solutions such as \cite{li_federated_2022} that utilize GANs may produce low-quality and non-diverse synthetic data, as GANs are known to suffer from instabilities such as mode collapse during training.
 
Considering the limitations above, we propose a novel and more effective data augmentation process in FL that uses foundation models to generate differentially private synthetic data. To the best of our knowledge, this is the first work that employs pre-trained foundation models to generate differentially private synthetic data to tackle the problem of Non-IID data in FL. Thus, our contributions are as follows:
\begin{itemize}
    \item We propose Differentially Private Synthetic Data Aided Federated Learning Using Foundation Models (DPSDA-FL) -- a new data augmentation strategy to enhance the performance of FL with Non-IID data -- and show the effectiveness of utilizing differentially private synthetic data generated from foundation models in cross-silo horizontal FL.
    \item We conduct experiments and evaluations on the CIFAR-10 dataset and observe an increase in the recall of the global model by up to 26\% and an accuracy enhancement of 9\%, demonstrating the efficacy of our approach over the baselines. We also provide an analysis of our results to guide further research. 
    % \textcolor{blue}{[MM: Be more concrete here; use the same numbers as you would use in the abstract.]} 
\end{itemize}

The remaining part of the paper is organised as follows. Section~\ref{sec:related_work} discusses related work. Section~\ref{sec:proposed_idea} introduces our methodology and proposes a new data augmentation strategy. Section~\ref{sec:evaluation} presents our experimental results and evaluations. Finally, Section~\ref{sec:conclusions} concludes the paper with a summary and an outline for future work.

\section{Background and Related Work}
\label{sec:related_work}

% \textcolor{blue}{[MM: I added `background` in the section title as you also give a bit of background about these concepts.]}

% \begin{figure*}[t]
%     \centering
%     \label{fig:arhitecture}
%     \includegraphics[width=1.0\textwidth]{LatextIJCAIImage.drawio.png}
%         \caption{Synthetic Data Aided Federated Learning Using Foundation Models.} 
%         %\textcolor{blue}{MM: Increase the font size of the text. It is unreadable.}} 
%         \label{fig:arhitecture}
% \end{figure*}

\begin{figure*}[t]
    \centering
    
    \includegraphics[width=1.0\textwidth]{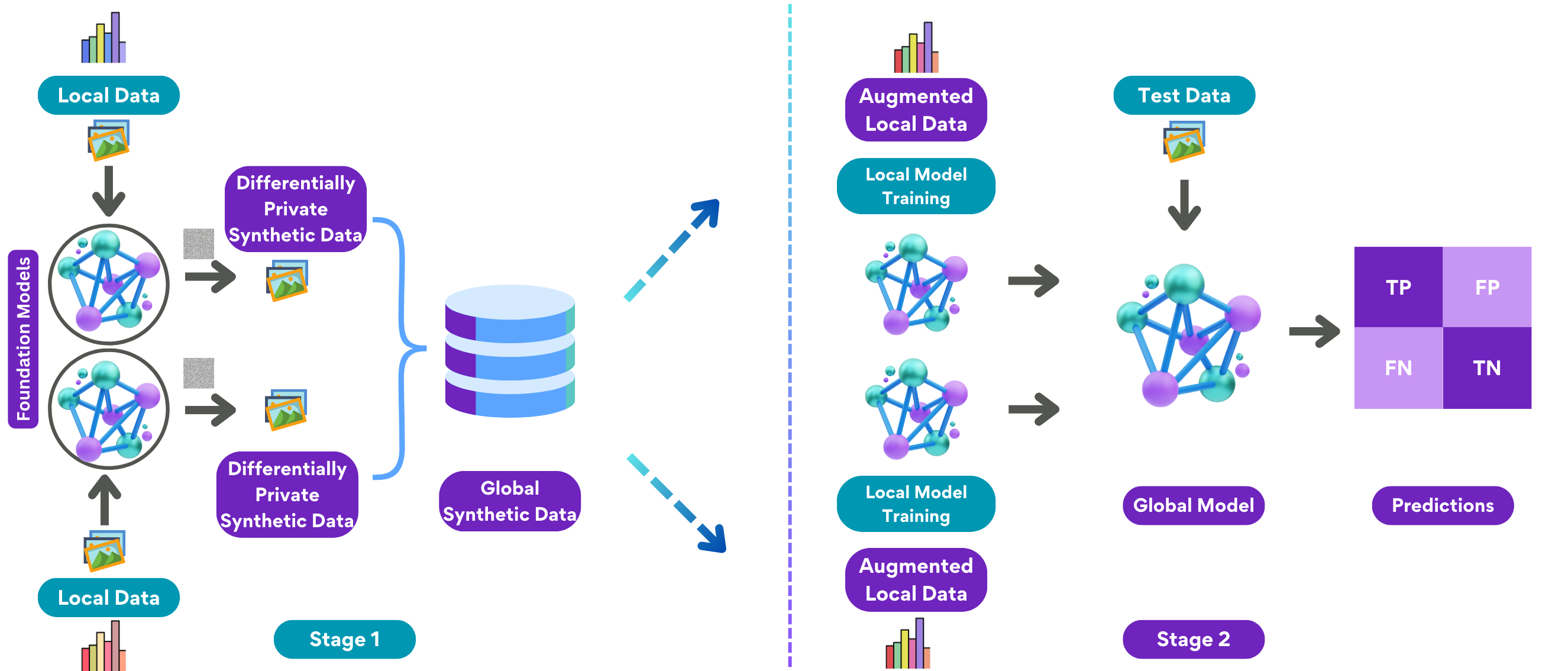}
        \caption{Differentially Private Synthetic Data Aided Federated Learning Using Foundation Models.} 
        %\textcolor{blue}{MM: Increase the font size of the text. It is unreadable.}} 
        \label{fig:architecture}
\end{figure*}

\subsection{Data Heterogeneity}

Data heterogenity is the degree of diversity in the datasets held by clients participating in FL. Data heterogeneity in FL arises from the differences in data distribution, data quality, and data quantity among participants. It can manifest in various forms in FL. Quantity Skew results from the differences in the amount of data held by clients, while Label Skew results from the differences in the classes of data held by individual clients \cite{qu_experimental_2021}. 

Several techniques have been proposed to address the challenge of data heterogeneity in FL. FedProx \cite{li_federated_2020} addresses heterogeneity in FL by integrating a proximal term into the training process. The proximal term leads to a reduction in the divergence of the local models from the global model by serving as a penalization term for the loss function of the local models. \cite{karimireddy_scaffold_2021} developed the stochastic controlled averaging algorithm, a modification of the federated averaging, which incorporates variance reduction to stabilize the local model towards the global model. However, these techniques are not effective in extreme cases of data heterogeneity. 
Another line of work uses GANs to mitigate the effects of data heterogeneity by generating additional training data for local data augmentation ~\cite{razavi-far_generative_2022} of which our method aligns with. However, despite using a similar approach to the GAN-based data augmentation methods, we locally generate differentially private synthetic data using foundation models to mitigate the effects of data heterogeneity, our solution generates more diverse synthetic data that is of higher quality than the GAN-based approaches. 

% \textcolor{blue}{[MM: Add a sentence summarising the limitation of these works that you solve.]}
% Our work aligns with this approach as we locally generate differentially private synthetic data using foundation models to mitigate the effects of data heterogeneity. 
% \textcolor{blue}{[MM: It is OK to say that your work is similar to x/y/z as you take the same approach/etc., however, it is crucial to highlight the difference as well. Paraphrase the sentence along these lines: `Although we use the same approach as in X as we locally generate differentially private synthetic data using foundation models to mitigate the effects of data heterogeneity, our solution [say what is different (or better say how your work is better than theirs.)]`]}

\subsection{Generative Adversarial Networks}

\textit{Generative adversarial networks (GANs)} are deep learning models comprising of two networks: the generator and discriminator. The generator produces synthetic data mimicking real data, challenging the discriminator to distinguish between them \cite{goodfellow_generative_2014}. Synthetic data from GANs share the statistical distribution of real datasets and, as such, can be used for dataset augmentation, enhancing model performance \cite{antoniou_data_2018}. 

\cite{zhang_feddpgan_2021} trained a GAN at the server side using FL and then shared the synthetic data across clients to improve the performance of FL. \cite{li_federated_2022} proposed Synthetic Data Aided Federated Learning (SDA-FL), where all clients receive a portion of locally synthetically generated data that is globally shared by the server. Despite the effectiveness of GAN-based methods in combating data heterogeneity problems in FL and enhancing the performance of the global model, these works have limitations. The instability of training GANs can result in low-quality synthetic samples with low utility \cite{azizi_synthetic_2023}. 

Recent works have addressed the underperformance of GANs in generating high-quality synthetic data by adopting diffusion models. Diffusion models are generative deep learning architectures that generate synthetic data by iteratively adding noise to real data and then removing this noise through a reverse diffusion process \cite{yang_diffusion_2024}. Diffusion models have been shown to produce high-quality data for computer vision applications~\cite{dhariwal_diffusion_2021,azizi_synthetic_2023}.  Diffusion models, however, can be challenging to train due to their high computational requirements, which are often beyond the reach of many. However, the emergence of foundation models has made access to pre-trained diffusion models more accessible. 

\subsection{Foundation Models}

Foundation models are a class of generative AI trained on large-scale data and can be modified to undertake various tasks with high precision \cite{zhou_comprehensive_2023}. Foundation models like Open AI's Stable Diffusion and DALL.E~\cite{ramesh_hierarchical_2022}  have become widely accessible. These pre-trained models can be used to generate high-utility synthetic data.

\subsubsection{Differentially Private Synthetic Data}

Synthetic data has been demonstrated to inadvertently reveal sensitive information about the original dataset it was generated from \cite{giomi_unified_2022}. Consequently, integrating privacy-preserving techniques into the synthetic data generation process is imperative. Differential Privacy (DP) is a method that introduces randomness while computing statistics to maintain the privacy of the underlying information \cite{dwork_algorithmic_2013}. It has emerged as the standard approach for enhancing the privacy of synthetic data due to its ability to offer provable privacy guarantees. Consequently, diffusion models can be trained using DP to safeguard the privacy of the synthetic data they produce. 

In \cite{ghalebikesabi_differentially_2023}, by fine\-tuning pre-trained diffusion models with tens of millions of parameters, high utility data with low Fréchet inception distance were generated privately. The synthetic data was employed for a downstream classification task, and state-of-the-art results were attained. A more recent method, PRIVIMAGE~\cite{li_privimage_2024}, generates differentially private synthetic images using foundation models by strategically selecting pre-training data. While this approach is effective, it incurs significant memory and time overheads. Another notable technique is Private Evolution (PE) \cite{lin_differentially_2024}, an algorithm that fine-tunes pre-trained diffusion models to generate synthetic data from private datasets while maintaining differential privacy. PE has demonstrated state-of-the-art results in image synthesis and requires no pre-training. In this study, we leverage PE to generate synthetic data for data augmentation.

\section{DPSDA-FL: Differentially Private Synthetic Data Aided Federated Learning Using Foundation Models}
\label{sec:proposed_idea}

This section proposes our novel technique, DPSDA-FL, that generates differentially private synthetic data for FL using foundation models. Figure~\ref{fig:architecture} gives a high-level overview of our proposal. DPSDA-FL works in two main stages, Stage 1 in which each Cross-Silo FL client uses a foundation model to locally generate differentially private synthetic data from their private data and then share part of the synthetic data with the central server to form a global synthetic data which will be utilized in Stage 2. In the next stage, the server distributes the global synthetic data to clients in order to enable them augment their local data with the diverse and high quality synthetic data. This augmentation leads to a less heterogeneous local data distribution by allowing clients to possess synthetic data from classes they do not possess or classes they possess a very limited sample from. This subsequently leads to a  more stable local model training that enhances the performance of the global model both in terms of its recall capability and its accuracy.
A more detailed overview of how DPSDA-FL works is presented below:

\begin{enumerate}
    \item \textbf{Unique label count information sharing:} At the start of the training process, clients share their unique label counts with the server to form a globally unique label count. This information will be used to share the synthetic global data with clients to ensure each client receives differentially private synthetic data from the classes they are deficient in.

\item \textbf{Local clients' synthetic data generation using foundation models:} 
To generate our differentially private synthetic data, we utilize the image-guided diffusion model DPSDA \cite{lin_differentially_2024}, as our foundation model. DPSDA is based on improved diffusion \cite{dhariwal_diffusion_2021}. To ensure the privacy of local training data, a local copy of the diffusion model is downloaded and hosted locally on client's devices. It mitigates privacy risks associated with diffusion models memorizing their training data, as shown in \cite{carlini_extracting_2023}. The local synthetic data \( D_{csyn} \) are then shared with the server to construct the global synthetic data \( D_{Gsyn} \).

\item \textbf{Global synthetic data distribution:}
The differentially private synthetic data from the previous step is then shared by the server with the local clients. The local data class information possessed by the server guides effective distribution, so each client only receives data from classes it lacks.

\item \textbf{Local data augmentation:} Clients then utilize the received synthetic data to augment their local data and homogenize the local data distribution. These synthetic data are of high quality and can enhance local model training.

\item \textbf{Federated training:}
With more stable local training aided by the augmented local datasets at each client's side, clients proceed to train a federated global model jointly. Note that DPSDA~\cite{lin_differentially_2024} does not necessitate any pre-training to generate the synthetic data, and the clients are assumed to be health institutions that can afford reasonable computational resources.

\end{enumerate}

% \textcolor{blue}{[MM: I do not see any different colours.]}

\begin{algorithm}[t]
\caption{DPSDA-FL}\label{Alg1}
\begin{algorithmic}[1]
    \STATE \textbf{Input Parameters:}
    \STATE $N$: Number of clients.
    \STATE $T$: Total number of rounds.
    \STATE $\alpha$: Learning rate.
    \STATE $w_t$: Initial model parameters.
    \STATE $w_{t+1}$: Updated model parameters.
    \vspace{0.5em}
\STATE \textbf{Initialization}
\STATE Clients share their unique label counts with the server
\STATE Clients generate DP synthetic data using Foundation Models
\FOR{$i = 1$ to $N$}
    \STATE Generate $D_{\text{syn}}^i$ from $D_c^i$
    \STATE Send $D_{\text{syn}}^i$ to server
\ENDFOR
\STATE Server forms global $D_{\text{Gsyn}}$ from $D_{\text{syn}}^i$
\STATE Distribute $D_{\text{Gsyn}}$ using unique label count
\FOR{$t = 1$ to $T$}
    \STATE Send $w_t$ to all clients
    \FOR{$i = 1$ to $N$}
        \STATE Augment $D_c^i$ with $D_{\text{Gsyn}}$
        \STATE Train model $L_i$ to update $w_{t+1}^i$
\STATE Server Initializes $w_0$
        \STATE Send $w_{t+1}^i$ to server
    \ENDFOR
    \STATE Aggregate $w_{t+1} = \frac{1}{N} \sum_{i=1}^{N} w_{t+1}^i$
\ENDFOR
\STATE Repeat until convergence
\end{algorithmic}
\end{algorithm}

Algorithm \ref{Alg1} outlines the pseudocode for DPSDA-FL. We consider a FL setting with a single semi-trusted central server \( S \) and \( N \) clients denoted by \( \{ C_{1}, C_{2}, ...C_{N}\} \). A horizontal FL setup is one where the data across the FL clients is partitioned horizontally, and clients share similar feature sets but different sample spaces. Each client possesses a local dataset \( D_{c} \), which is a subset of the global dataset \( D_{G} \). \( D_{G} \) follows a normal distribution and consists of \( k \) classes of data.  However, \( {D_c} \) does not follow a normal distribution as the data distribution amongst the clients is Non-IID. Some clients may possess fewer samples than others, leading to quantity skew or some classes of data but not others resulting in label skew. As we are considering a cross-silo FL setup, all the clients participate in training rounds, and each local model \(L_n\) contributes to the global model aggregation. The objective is to produce a single global model \(G\) that performs well on the global test data.

\begin{table}[t]
\centering
\caption{\textbf{Experimental Settings}}
\label{table:exp-settings}
\begin{tabular}{l|p{0.25\textwidth}}
% \begin{tabular}{ll}
\toprule
\textbf{Name} & \textbf{Value} \\
\midrule
FL architecture & Cross-Silo Horizontal FL \\
Dataset & CIFAR-10 \\
NN architecture & CNN \\
Number of clients & 5 \\
Number of local epochs & 2 \\
Number of global rounds & 20 \\
Learning rate & 0.1 \\
Batch size & 32 \\
Optimizer & Stochastic gradient descent \\

\bottomrule
\end{tabular}
\end{table}

\section{Experiments and Evaluations}
\label{sec:evaluation}

% \textcolor{blue}{[MM: There are too many subsections. There should be no more than 3-4 such subsections. I grouped them.]}

\subsection{Experimental Setting}

Below we describe the experimental settings used in our evaluation. These settings are also summarised in Table~\ref{table:exp-settings}.

\subsubsection{Dataset}
We performed our experiments on the CIFAR-10 dataset, which is a benchmark dataset used for image recognition. It consists of 50,000 training samples and 10,000 testing samples. The dataset is mostly utilized to evaluate the classification accuracy of Convolutional Neural Networks (CNN). We used the entire 10,000 images to test the accuracy of the global model for our approach and the baselines.

\subsubsection{Differentially Private Synthetic Dataset}
We deployed five pre-trained diffusion models to generate synthetic local data for each client. To limit privacy risks associated with the honest but curious server, we assumed each client only generated and shared at most 50\% of its number of classes. We generated 5000 differentially private synthetic images for each class of the CIFAR-10 dataset and selected a subset to be used for augmentation. The generated images were 64 x 64; as such, they were resized to 32 x 32, which is the original size of the CIFAR-10 images.

\subsubsection{Data Distribution}
We evaluated the effectiveness of our approach by simulating real-world FL participants with varying data distributions that follow a Non-IID fashion. In other words, the local data distribution of each client is not representative of the global dataset. To simulate extreme label skew for our experiments, we followed the work~\cite{li_federated_2020}; each client received samples from only two classes.  
% While additionally, clients received variable number of samples following the power law for quantity skew.

% \subsubsection{Quantity Skew}

% \subsubsection{Label Skew}

\begin{table*}
    \centering
    \caption {{Classification Accuracy of the Global Model in FedAvg and FedProx Compared with DPSDA-FL with 5 Clients}}
        \label{tab: Global Model1}
    \begin{tabular}{ccccc}
        \toprule
        Approach  & Data Augmentation  & \% of Synthetic Data Shared & Classes/Client & Global Model Accuracy  \\
        \midrule
         FedAvg     & No & $0$   & $2$   &  $28.30\pm 2.20\% $     \\
        FedProx         & No & $0$ & $2$     &  $31.70\pm 2.26\% $       \\
        DPSDA-FL    & Yes & $50$  & $2$      &  $37.20\pm 0.44\% $      \\
        % FedProx+ DPSDA-FL  & Yes & 0 & 0     &  \%         \\
        % Case 5          & Non-IID  & Yes & 0  & 0 & 0     &  \%        \\
        \bottomrule
    \end{tabular}
\end{table*}

\begin{table*}
    \centering
    \caption {{Recall of the Global Model in FedAvg and FedProxCompared with DPSDA-FL with 5 Clients}}
    \begin{tabular}{cccccc}
        \toprule
        Approach & Recall of the Plane Class & Recall of the Cat Class & Recall of the Ship & Class Recall of the Truck Class \\
        \midrule
        FedAvg    & 55.9\%  & 15.6\%  & 35.1\%   &  46.5\%      \\
        FedProx   & 40.6\%  & 19.3\% &  53.2\%    &  44.6\%         \\
        DPSDA-FL  & 58.86\% & 42.4\% & 58.6\%       &  56.46.0\%         \\
        
        \bottomrule
    \end{tabular}
    \label{tab: Global Model2}
\end{table*}

\subsubsection{Neural Network Architecture}
We used CNN for all our experiments. The CNN includes two convolutional layers, each followed by a ReLU activation and a max-pooling layer. It also includes two fully connected layers followed by a ReLU activation. The final output layer uses a log-softmax activation to produce class probabilities. We used stochastic gradient descent (SGD) and negative log-likelihood as our local model optimizer and loss function.

\subsubsection{Cross-Silo Horizontal FL}
In cross-silo horizontal FL (HFL), clients are usually smaller in number compared to cross-device horizontal FL, where there are many devices collaboratively training a model. Also, unlike cross-device HFL, where a fraction of clients are selected to participate in each round, in cross-silo HFL, all the clients usually participate in all communication rounds. Cross-silo setups are also traditionally made up of larger institutions with abundant computational resources serving as clients compared to cross-device, where there are typically resource-constrained smaller devices participating in the training process. 

\subsubsection{Evaluation Metrics}
We used the top-1 accuracy of the global model to evaluate how effectively our data augmentation technique can enhance FL. We also measured and compared the recall of the global model produced by the baselines and our approach. Recall is an essential metric for machine learning models that are deployed in safety-critical sectors such as healthcare, where FL is a good candidate. This is because false positives in such domains entail significant consequences. We reported the mean and standard deviation of the global model accuracy from running each experiment three times. We also reported the mean of the recall.

\subsubsection{Baselines}
\textbf{{Federated Averaging}}: To compare the effectiveness of our approach, we used federated averaging \cite{mcmahan_communication-efcient_2017}, which is the vanilla version of FL where clients share their local model parameters with the central server, which is then aggregated to form an updated global model. \\
\textbf{{Federated Optimization (FedProx)}}: We implemented FedProx \cite{li_federated_2020} with the same number of clients as in our work and used the value of 0.001 for mu, which is the proximal term that aids in combating the effect of data heterogeneity in FL with Non-IID Data.

% \subsection{Hyperparameter Settings} For all the experiments we used the following the parameters: Local Batch size = 32, Number of Local Epoch = 2, and Number of Global Rounds = 10.

%\begin{figure}[!htb]
%\centering
%\minipage{0.50\textwidth}
%  \includegraphics[width=\linewidth]{IID_CM Exp4.png}
%  \caption{Global Model Trained on IID Data Using FedAvg}\label{fig:FedAvgIID}
%\endminipage\hfill
%\minipage{0.50\textwidth}
%  \includegraphics[width=\linewidth]{FedAvg Exp4.png}
%  \caption{Global Model Trained on Non IID Data Using FedAvg}\label{fig:FedAvg}
%\endminipage\hfill
%\minipage{0.50\textwidth}
%  \includegraphics[width=\linewidth]{FedAvg Exp4.png}
%  \caption{Global Model Trained on Non IID Data Using FedProx}\label{fig:FedProx}
%\endminipage\hfill
%\minipage{0.50\textwidth}%
%  \includegraphics[width=\linewidth]{DPSDA CM_Exp4.png}
%  \caption{Global Model Trained on Non IID Data Using DPSDA FL}\label{fig:DPSDA - FL}
%\endminipage
%\end{figure}

\begin{figure*}[t!]
    \centering
    \begin{subfigure}[t]{0.49\textwidth}
        \centering
  \includegraphics[width=\linewidth]{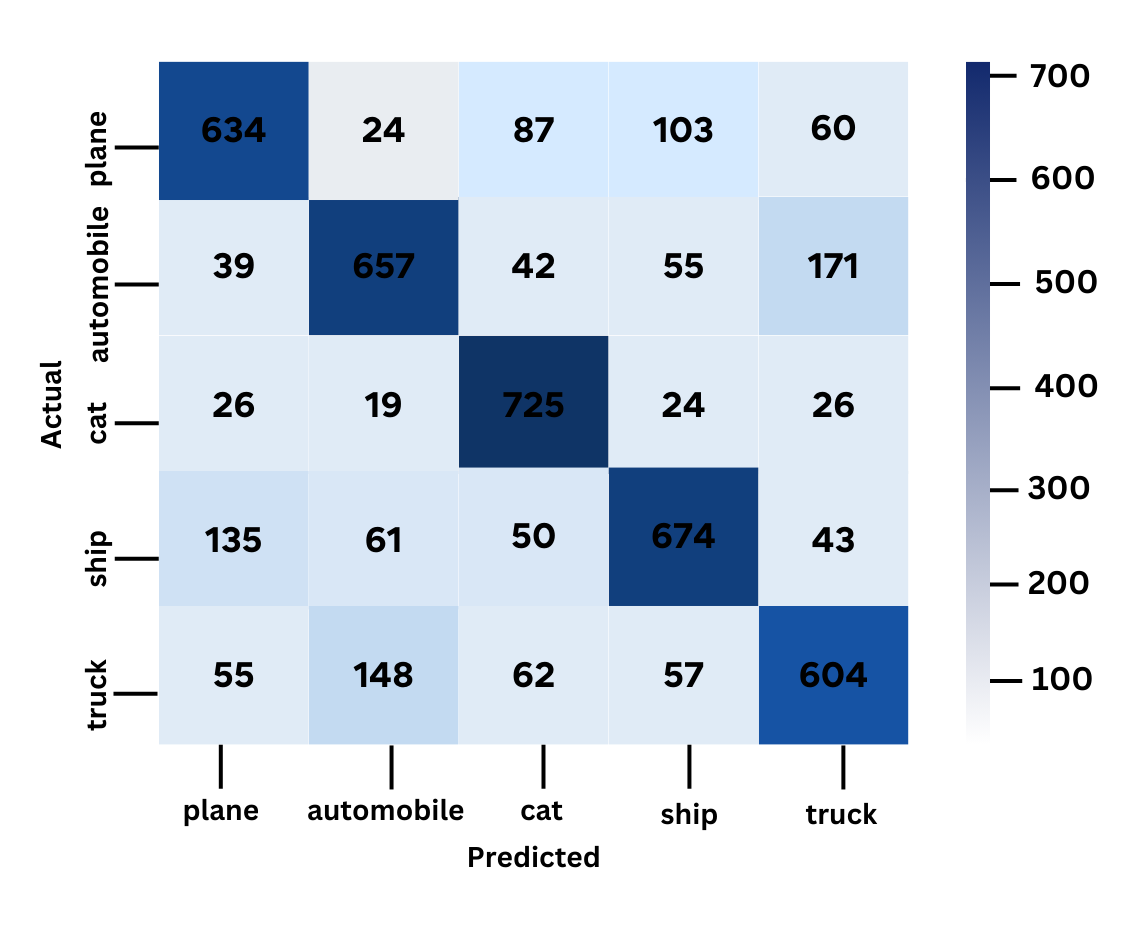}
  \caption{Global Model Trained on IID Data Using FedAvg}\label{fig:FedAvgIID}
    \end{subfigure}
    \hfill
    \begin{subfigure}[t]{0.49\textwidth}
        \centering
  \includegraphics[width=\linewidth]{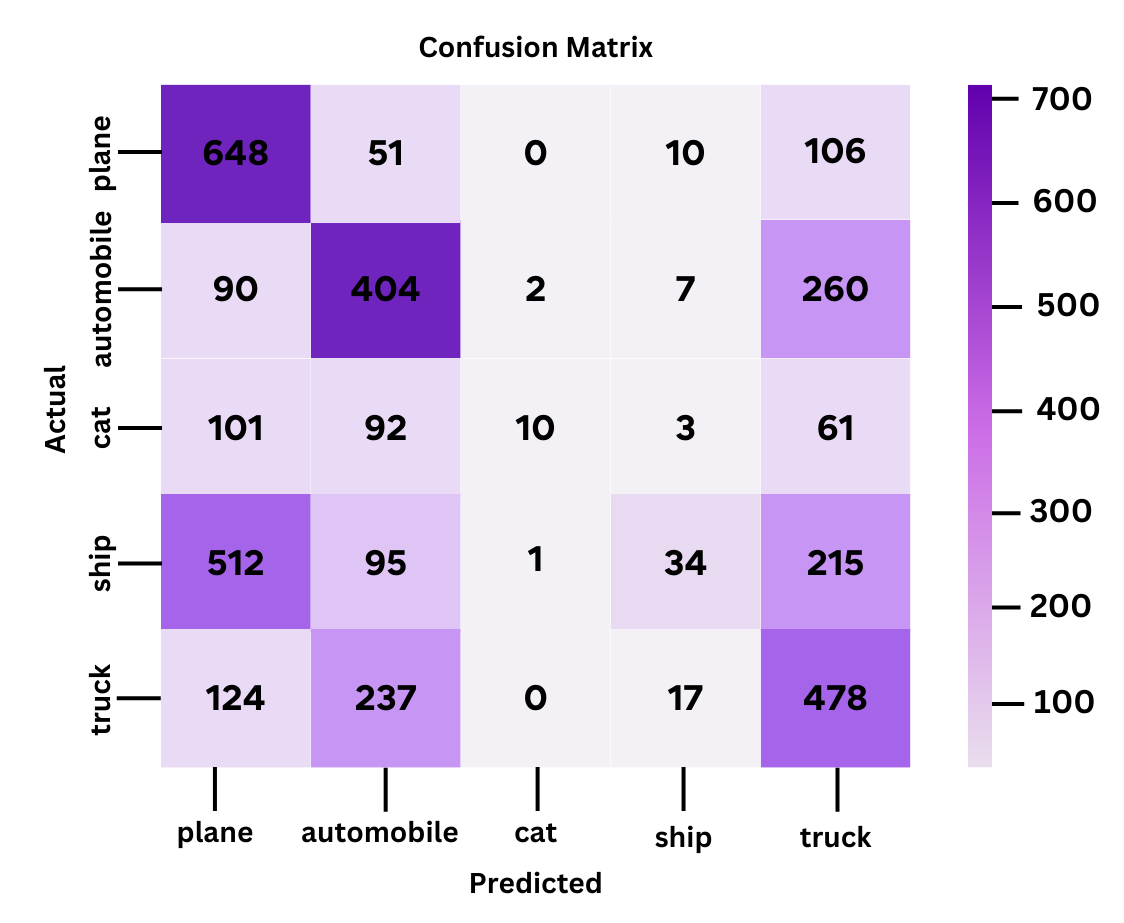}
  \caption{Global Model Trained on Non-IID Data Using FedAvg}\label{fig:FedAvg}
    \end{subfigure}
    \hfill
    \begin{subfigure}[t]{0.49\textwidth}
        \centering
  \includegraphics[width=\linewidth]{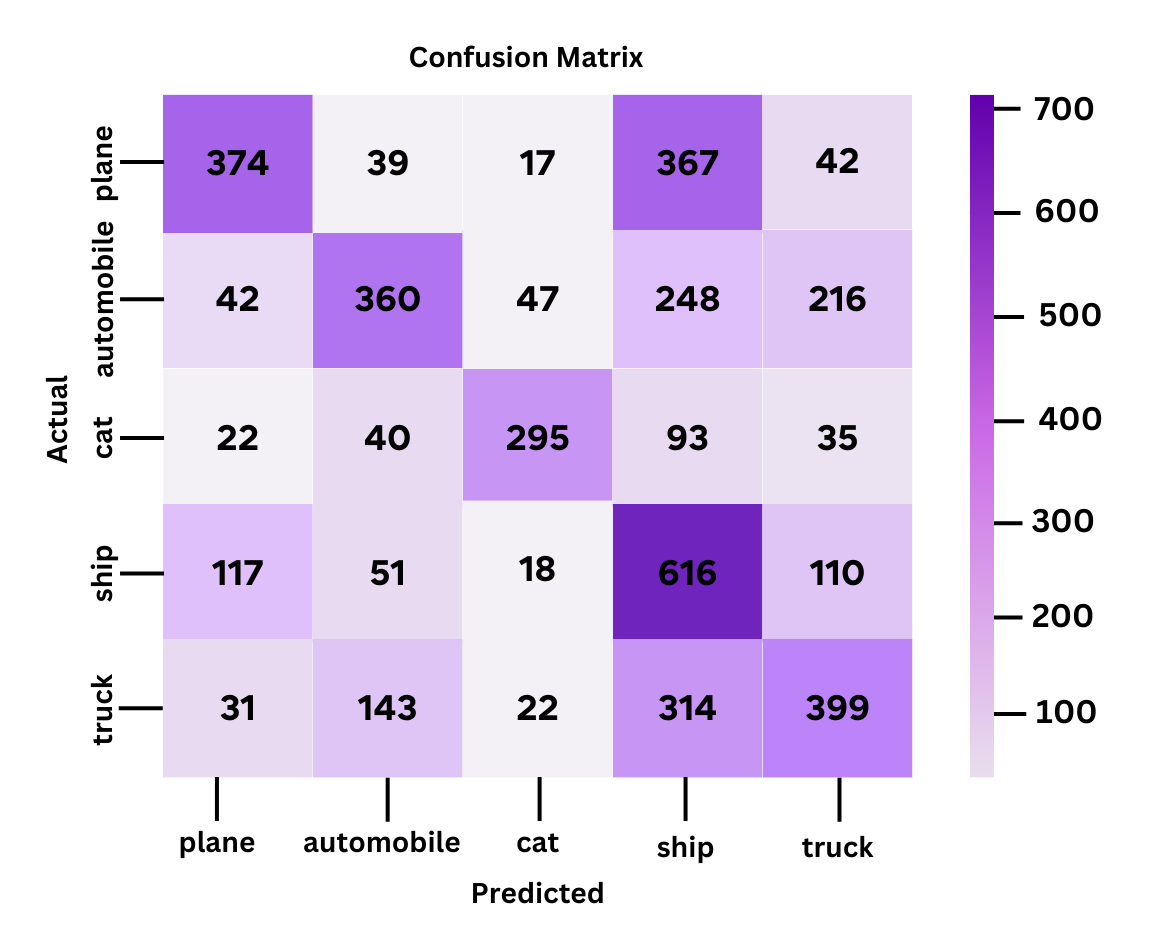}
  \caption{Global Model Trained on Non-IID Data Using FedProx}\label{fig:FedProx}
    \end{subfigure}
        \hfill
    \begin{subfigure}[t]{0.49\textwidth}
        \centering
  \includegraphics[width=\linewidth]{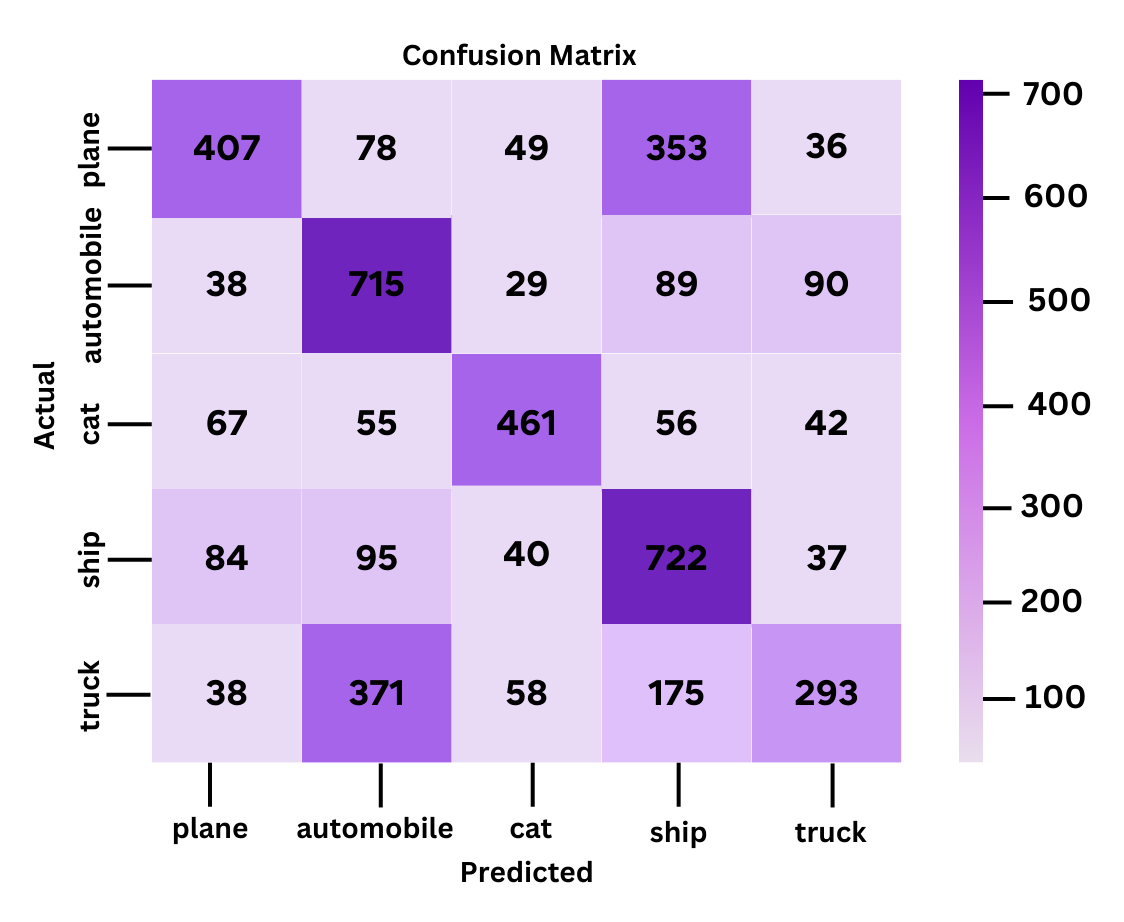}
  \caption{Global Model Trained on Non-IID Data Using DPSDA FL}\label{fig:DPSDA - FL}
    \end{subfigure}
    \caption{\centering{{Confusion matrices highlighting the correct predictions and misclassifications made by the various approaches.}}}
    \label{fig:confusion-matrixes}
\end{figure*}

\subsection{Results and Discussion}
Our experimental results are summarised in Tables~\ref{tab: Global Model1} and \ref{tab: Global Model2}, as well as in Figure~\ref{fig:confusion-matrixes}. We visualized our models' performances using confusion matrices as they provide a clear insight into the model's ability to make correct predictions for both positive and negative cases. The darker shades of colour in each matrix represent these correct predictions.
Our findings reveal a promising outcome. For the global FL model trained on IID Data using FedAvg, which represents the most ideal case, as depicted in {Figure~\ref{fig:FedAvgIID}}, the majority of classes were correctly identified, as evidenced by a darkened diagonal. In contrast, the global model trained on Non-IID using FedAvg, where each client possesses data from only 2 classes, struggles to correctly identify positive and negative cases, as shown in {Figure~\ref{fig:FedAvg}}. This struggle highlights the challenge faced by models in such scenarios. The global model trained using FedProx {Figure~\ref{fig:FedProx}} shows a better performance than FedAvg, due to the proximal term added to the loss function of local clients to prevent significant divergence. However, DPSDA-FL demonstrated an even more accurate and enhanced global model, as shown in {Table \ref{tab: Global Model1}}  and {Figure~\ref{fig:DPSDA - FL}}. This can be attributed to the more stable local training for clients aided by the differentially private synthetic data. The recall of the classes for which the differentially private synthetic data generated by the foundation models is shared with other clients tends to improve as well, as shown in Table \ref{tab: Global Model2}. This suggests that the local models that make up the global model were able to effectively learn features of those classes from the high utility and diverse synthetic data. As such, the global model is able to identify more data samples from those classes and also distinguish the classes more accurately than others, compared to FedAvg and FedProx. 

% \textcolor{blue}{[MM: It seems that you call your approach DPSDA-FL, but you use this term only now, in the end of the paper. I thought it was another state-of-the-art techniqe. I got confused. Please introduce this name already in the introduction when you state the contributions, then keep using it in the next sections too instead of `our methodology/approach/strategy/etc.`]} 

\section{Conclusion}
\label{sec:conclusions}

In this paper, we present a new data augmentation strategy that has the potential to significantly enhance the performance of cross-silo horizontal FL with Non-IID. By mitigating the effect of data heterogeneity, DPSDA-FL, which utilizes differentially private synthetic data generated by pre-trained foundation models, can improve the local training and convergence of the global model. Our experimental results demonstrate that DPSDA-FL can effectively improve the class recall and classification accuracy of the global model in FL with Non-IID issues. %, thereby making a significant contribution to the field of federated learning.

We plan on evaluating our approach by conducting more experiments and generating local synthetic data using a limited number of private data samples. %, as that may be the case in FL with Non-IID, where all the clients might possess minimal data from a particular class. 
We also leave experimenting with other datasets that do not overlap with the training datasets of the foundation models for future work. \\

\section*{Acknowledgements}

This work was supported by The University of Manchester and EPSRC through the EnnCore project [EP/T026995/1].

%\newpage
\bibliographystyle{named}
\bibliography{ijcai24}

\end{document}